\documentclass{llncs}
\usepackage{amssymb,amsfonts,amsmath}

\usepackage{graphicx}
\usepackage{color}
\usepackage[ruled]{algorithm2e}

\usepackage{footnote}
\usepackage{float}

\makeatletter
\newcommand*{\defeq}{\mathrel{\rlap{%
                     \raisebox{0.3ex}{$\m@th\cdot$}}%
                     \raisebox{-0.3ex}{$\m@th\cdot$}}%
                     =}
\makeatother

\begin{document}


\title{Automated Unsupervised Segmentation of Liver Lesions in CT scans via Cahn-Hilliard Phase Separation }


\author{
Jana Lipkov\'a \inst{1}, Markus Rempfler \inst{1}, Patrick Christ \inst{1}, John Lowengrub\inst{2}, Bjoern H. Menze\inst{1}
\institute{
Department of Informatics \& Institute for Advanced Study, Technical University of Munich, Germany
\and
Departments of Mathematics, \& Center for Complex Biological Systems \& Chao Family Comprehensive Cancer Center, University of California, Irvine, USA
}
\email{jana.lipkova@tum.de}
}


\maketitle

\begin{abstract}
The segmentation of liver lesions is crucial for detection, diagnosis and monitoring progression of liver cancer. However, design of accurate automated methods remains challenging due to high noise in CT scans, low contrast between liver and lesions, as well as large lesion variability. We propose a 3D automatic, unsupervised method for liver lesions segmentation using a phase separation approach. It is assumed that liver is a mixture of two phases: healthy liver and lesions, represented by different image intensities polluted by noise. The Cahn-Hilliard equation is used to remove the noise and separate the mixture into two distinct phases with well-defined interfaces. This simplifies the lesion detection and segmentation task drastically and enables to segment liver lesions by thresholding the Cahn-Hilliard solution. The method was tested on 3Dircadb and LITS dataset.
\end{abstract}

\section{Introduction}\label{sec:Intro}
Liver is one of the most common cancer sites, including primary tumours like hepatocellular carcinoma and metastatic tumours that have spread from the breast, colon and prostate. Computer tomography (CT) is routinely used to detect and evaluate treatment response of liver lesions. In clinical practise, liver lesions are segmented by manual or semi-manual methods. However, these are time consuming and subjective, with an intra- and interobserver variability up to 11~\% in volume difference on liver CT scans \cite{Bellon:1997}. To overcome these difficulties, several semi-automated and automated methods were proposed. Semi-automated methods include support vector machine with affinity constrains propagation \cite{Freiman:2011}, hidden Markov fields  \cite{Hame:2012}, level set methods \cite{Li:2013}, sigmoid edge modelling \cite{Foruzan:2016} and mathematical morphology \cite{Belgherbi:2014}. Automated methods include k-means classification \cite{Massoptier:2008}, object-based image analysis \cite{Schwier:2011} and convolutional neural networks \cite{Christ:2017b}. An advantage of automated over semiautomated methods is their reproducibility, since they do not require human interactions. Despite significant efforts, the performance of automatic methods remains relatively poor, especially in comparison with segmentation methods for other lesion sites.  The main challenges of liver lesion segmentation include high levels of noise, low liver-lesion contrast and variations of image intensities  caused by different acquisition protocols, tissue abnormalities such as surgical resection, metal implants and changes due to treatment. For instance, the mean liver CT values of 3Dircadb \cite{Iircad} datasets vary by an order of magnitude, which complicates the use of intensity based methods. Furthermore, a significant variation in lesions shape and structure compromise efficiency of supervised methods. 
\\
We propose a novel automated unsupervised method for the enhancement and segmentation of hypointense lesions in liver CT scans via phase field separation. In chemistry, phase separation is a mechanism in which a mixture of two components separates into distinct phases with different chemical compositions. The Cahn-Hilliard equation is a partial differential equation that describes phase separation driven by gradients in chemical potentials \cite{Cahn:1958}. We consider liver CT as a mixture of two phases, healthy liver and lesions, represented by different image intensities. The Cahn-Hilliard equation is used to remove the noise and separate the mixture into two distinct phases with well defined interface separating the phases. The lesions are then segmented by thresholding the Cahn-Hilliard solution. This approach has several desirable properties: it is 3D, edge preserving and robust to noise, variation of intensities and lesions diversity. In comparison to other edge preserving smoothing methods, including bilateral and image guided filtering or anisotropic diffusion, phase separation is an energy minimisation problem which can be applied to data with different noise or image intensities.  
%
%
%
\section{Method}\label{sec:Method}
The Cahn-Hilliard equation describes the spatiotemporal evolution of phase separation in a mixed system. Let us assume a system with two phases, \textit{A} and \textit{B}. The state of the system at spacial location $(x,y,z) \in \mathbb{R}^{3}$ and time $t$ can be represented by a phase field function (pff) $\psi := \psi(x,y,z,t) \in [0,1]$, with $\psi = 1$ and $\psi=0$ indicating domains of the separated phases. The free energy of the system in a domain $\Omega$ can be modelled as \cite{Cahn:1958}
\begin{eqnarray}\label{eq:FreeEnergy}
E_{\varepsilon}(\psi) = \int_{\Omega} f(\psi) + \frac{\varepsilon^2}{2} \left| \nabla  \psi \right|^2 d\,V,
\end{eqnarray} 
where $f(\psi)$ is the bulk free-energy density in phases $A$ and $B$, $\varepsilon$ is the prescribed interface thickness and $\frac{\varepsilon^2}{2}|\nabla \psi |^2$ is the additional free-energy density at the interfaces between the phases. To ensure a separation in two distinguish phases, it is assumed that $f(\psi)$ is a double-well potential, which can be modelled as $f(\psi) = \frac{1}{4} \psi^2 (1-\psi)^2$. Then phase separation of the system driven by the difference in chemical potentials between the phases can be modelled by the Cahn-Hilliard equations:
\begin{eqnarray}\label{eq:CH}
\frac{\partial \psi}{\partial t} = \nabla \cdot \left( M(\psi) \nabla \mu \right), \quad \in \Omega,\\
\mu = \frac{\delta E_{\varepsilon}(\psi)}{\delta \psi}  := \frac{d f(\psi)}{d \psi} - \varepsilon^2 \Delta \psi,
\end{eqnarray}
where $\mu$ is the chemical potential of the system, defined as the variational derivation of the systems free energy. Taking the mobility term $M(\psi) = \sqrt{4 f(\psi)}$ inhibits long-range diffusion and tends to preserve the volumes of the individual lesions. The Cahn-Hilliard equation describes the evolution of a system with high energy, represented by mixed phases, to a system with lower energy characterised by the separated phases. In contrast to ill-posted anisotropic diffusion problem \cite{Guidotti:2015}, the existence and uniqueness of the Cahn-Hilliard solution is guaranteed by the existence of the free energy (Lyapunov) functional. Details on the derivation and properties of the Cahn-Hilliard equation can be found in \cite{Lee:2014}.
\\
The equation \eqref{eq:CH} is discretised by finite differences in space and forward Euler in time. It is implemented in a multi-resolution adapted grid solver, a 3D extension of the 2D solver presented in \cite{Rossinelli:2015}. The adaptive grid approach enables fast evaluation, with typical simulation time around 7 minutes per liver volume with 4 Opteron6174 cores.
\section{Experiments and results}\label{sec:Experiments}
The Cahn-Hilliard separation (CHS) method consists of three main steps: data preprocessing, phase separation and lesions segmentation. The workflow of the CHS method is depicted in Fig.\ref{fig:CHflow}.
\subsubsection{Data and data preprocessing:}
\begin{figure}[t]
\centering
\includegraphics[width=1\linewidth]{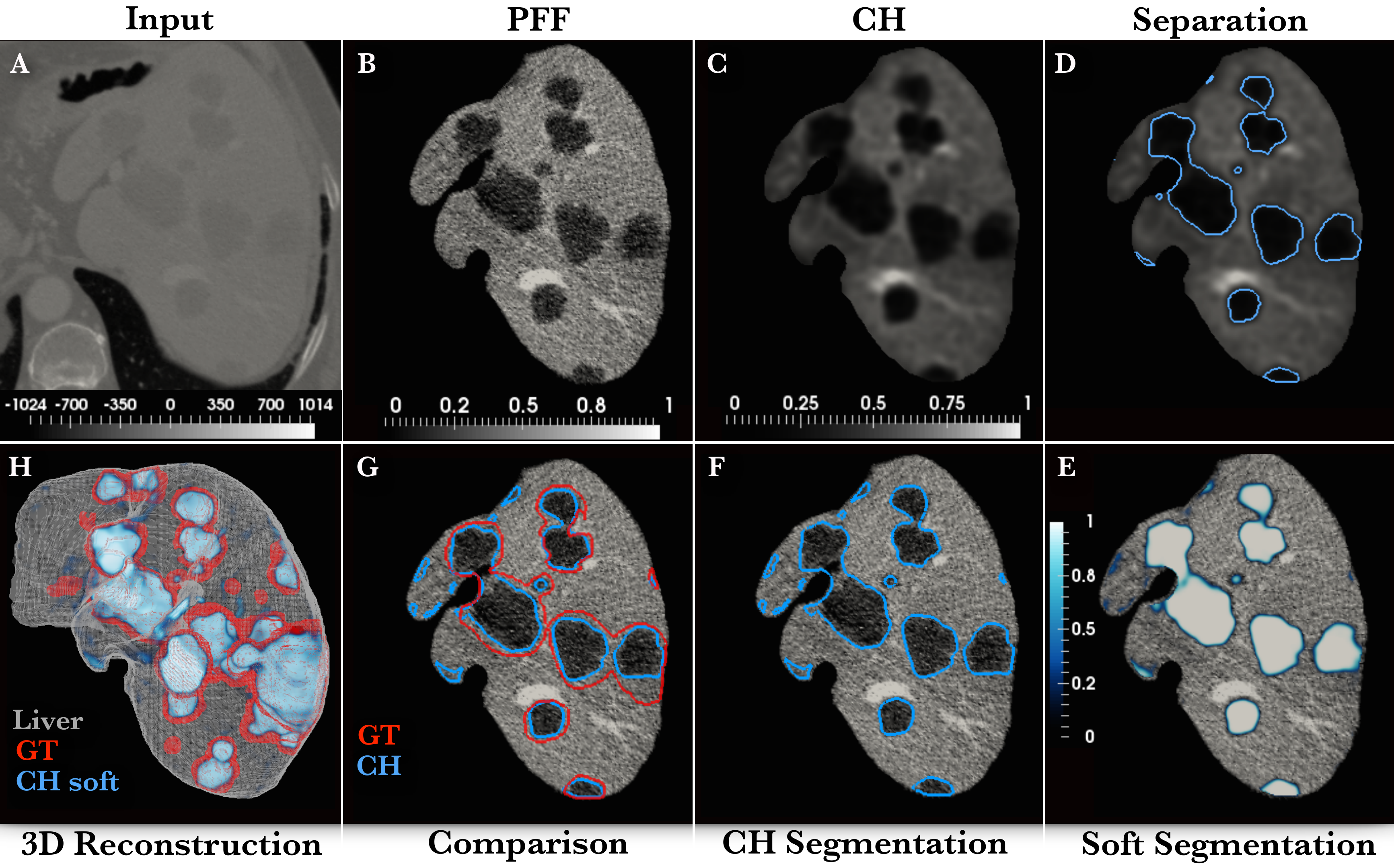}
\caption{A-H: A workflow of phase separation and lesions segmentation. A) cropped liver CT scan, converted into a phase field function (PFF) B), is used as initial condition for the Cahn-Hilliard (CH) equation, with the solution shown on C). D) the hard separation of the phases (blue line), is converted into a soft probabilistic segmentation E). F) final hard segmentation (blue) in comparison with the ground truth (GT) (red) F). H) comparison of the GT (red) and soft segmentation (white-blue colormap) in 3D representation.} \label{fig:CHflow}
\end{figure}

We conducted experiments on training dataset from the Liver Tumor Segmentation Challenge (LITS) \cite{LITS}, which also includes 3Dircadb data. The LITS datase contains abdominal CT scans acquired at various centres, with different acquisition protocols and resolution. All data are preprocessed in the following way. First, the abdominal CT scan is cropped into a box $\Omega$ containing a liver mask $\chi$, (Fig. \ref{fig:CHflow} A). Liver intensities are then clipped to a range $[0,200]$ HU, to exclude atypical liver values like metal implants. To account for specific image intensities, a $95\%$ credibility interval $[a,b]$  of the clipped liver CT is computed. The clipped liver CT is then clip one more time to the range $[0,b]$. Afterwards, the liver CT is normalised to take values in $[0,1]$ and this is used to initialise the pff $\psi$ within $\chi$ (Fig. \ref{fig:CHflow} B). To prevent border artefacts, the background of the liver CT is considered as a healthy liver. This is achieved by assigning a liver-like intensity to the phase field function outside the liver, i.e. it is assumed that $\psi = 0.55$ in $\Omega \setminus \chi$.
\begin{figure}[t]
\centering
\includegraphics[width=1\linewidth]{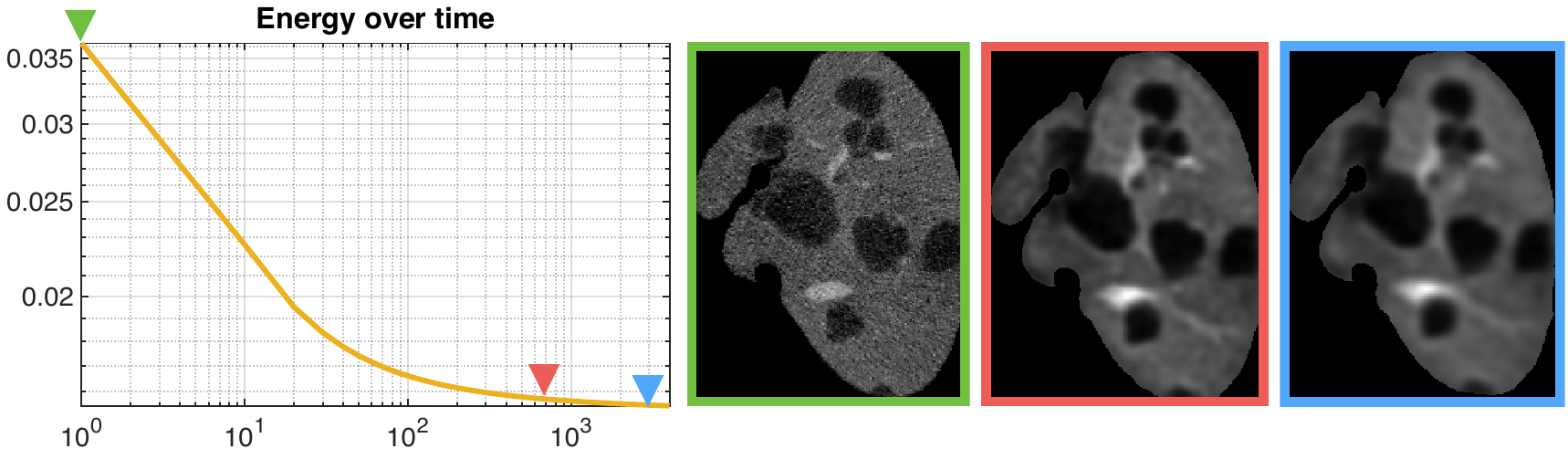}
\caption{ Log-log plot showing evolution of the system from high energy state, caused by mixed liver-lesion interface, to low energy state with separated phases.} \label{fig:Energy}
\end{figure}
\begin{figure}[t]
\centering
\includegraphics[width=1\linewidth]{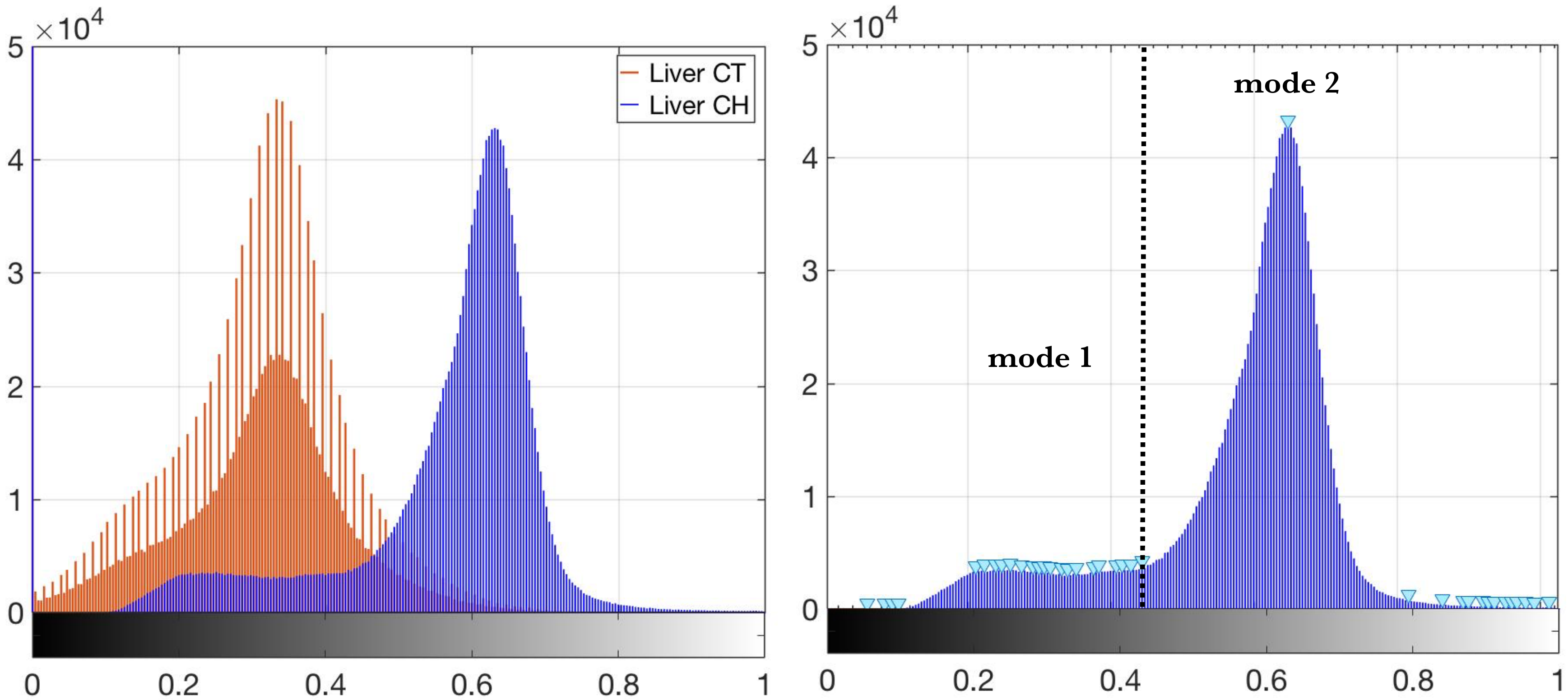}
\caption{Left: comparison of image intensity histogram (IIH) of the normalised liver CT before (red) and after the separation (blue). Right: IIH after the phase separation, with the detected local maxima indicated by the triangles. The dashed line indicates hard separation between lesions (mode 1) and liver (mode 2).} 
\label{fig:IIH}
\end{figure}
%
%
\subsubsection{Phase separation:}
The thickness of the liver-lesion interface $\varepsilon$ is set to 6 voxels, i.e 3 voxels of smoothening per phase, which is sufficient to smooth out noise but preserve small lesions. Eq. \eqref{eq:CH}, with $\psi$ defined above and no-flux boundary conditions on $\partial \Omega$ is evolved in time until the systems energy $E_{\varepsilon}$ approaches its minimum (Fig. \ref{fig:Energy}). In all tests, 700 times steps (iterations) are found to be sufficient to capture changes in the energy. Figure \ref{fig:Energy} shows, that the solution of the systems does not change significantly by evolving the system longer in time. The solution of the Cahn-Hilliard equation, with initial CT scan on Fig. \ref{fig:CHflow} A), is  shown on Fig. \ref{fig:CHflow} C). The phase separation dynamics removed noise and enhanced the liver-lesion contrast, while preserving the interface. The separation of the phases is also apparent from the image intensity histogram (IIH) of the normalised liver CT before and after the separation (Fig. \ref{fig:IIH}). The histogram is divided into 255 bins corresponding to the gray scale levels. The spikes visible on the original liver CT histogram are caused by anisotropic data resolution. After the CHS, the originally unimodal liver CT histogram separates into two modes, one for each phase, allowing lesions segmentation by histogram thresholding. In the case of binary system, the separation of the phases is give by $\psi = 0.5$. However, this is not the case for the heterogeneous liver scan.
\subsubsection{Lesions segmentation:}
Several methods have been proposed for automated histogram separation including the Otsu, Triangle and Isodata methods. However, these methods failed to detect the separation in the case of small lesions. Instead, we propose to compute the separation by detecting local maxima (peaks) of the IIH. The $i$-th element of the image histogram $iih(i)$, is defined as a peak, if $iih(i+1) - 2\,iih(i) + iih(i-1) < 0.$ Let \textbf{p} be a vector of the detected peak locations and $\textbf{I(p)}$ the corresponding image intensities. Let $p_j$ be the global maximum of $\textbf{p}$ with intensity $I(p_{j})$. Then the peak $p_{k}$ indicating separation between liver-lesion modes is identified by the Algorithm \ref{algorithm}. The while-loop in the algorithm ensures a correct histogram separation even if multiple peaks are detected within the liver mode. The separation of lesion (mode 1) and liver (mode 2) is shown in Fig. \ref{fig:IIH} (right) and the corresponding separation of the Cahn-Hilliard solution at the intensity $I_{0} = I(p_{k})$ is shown on Fig. \ref{fig:CHflow} D). However, a single iso-value $I_{0}$ might not be optimal for all lesions, especially small lesions might be under segmented. To overcome this issue, the hard interface separation $I_{0}$ can be translated into a soft probabilistic one as follows. The phase interface after CHS can be approximated by hyperbolic tangent
\begin{eqnarray}\label{eq:tanh}
\psi_{soft}(I) = \frac{1}{2}\left[ 1 + \tanh\left(  \frac{(I_{0} - I)}{2\sqrt{2}\varepsilon} \right)  \right]   = \frac{1}{1 + \exp\left(-\frac{(I_{0} - I)}{\sqrt{2} \varepsilon} \right)}.
\end{eqnarray} 
The soft probabilistic segmentation (Fig. \ref{fig:CHflow} E) is thresholded to $[0.15, 1]$ range to obtain the final segmentation (Fig. \ref{fig:CHflow} F). Figure \ref{fig:CHflow} (G and D) shows a comparison of CHS and the ground truth (GT) segmentation. The interface between liver and lesions is preserved in CHS, which leads to a better lesions delineation compare to the manual segmentation, which tends to over-segment some lesions.
\begin{algorithm}[t]\label{algorithm}
 \caption{\textsc{Histogram separation by detecting local maxima.}}
$k = j-1$\\
\While{  $I(p_{k})  > 0.75 \times I(p_{j})$  } {
$k=k-1$\;
}
\If{$(k=0)$}{
$I(p_k) \defeq 0$\;
}
\end{algorithm}
%
%
%
%
%
\begin{figure}[t]
\centering
\includegraphics[width=1.0\linewidth]{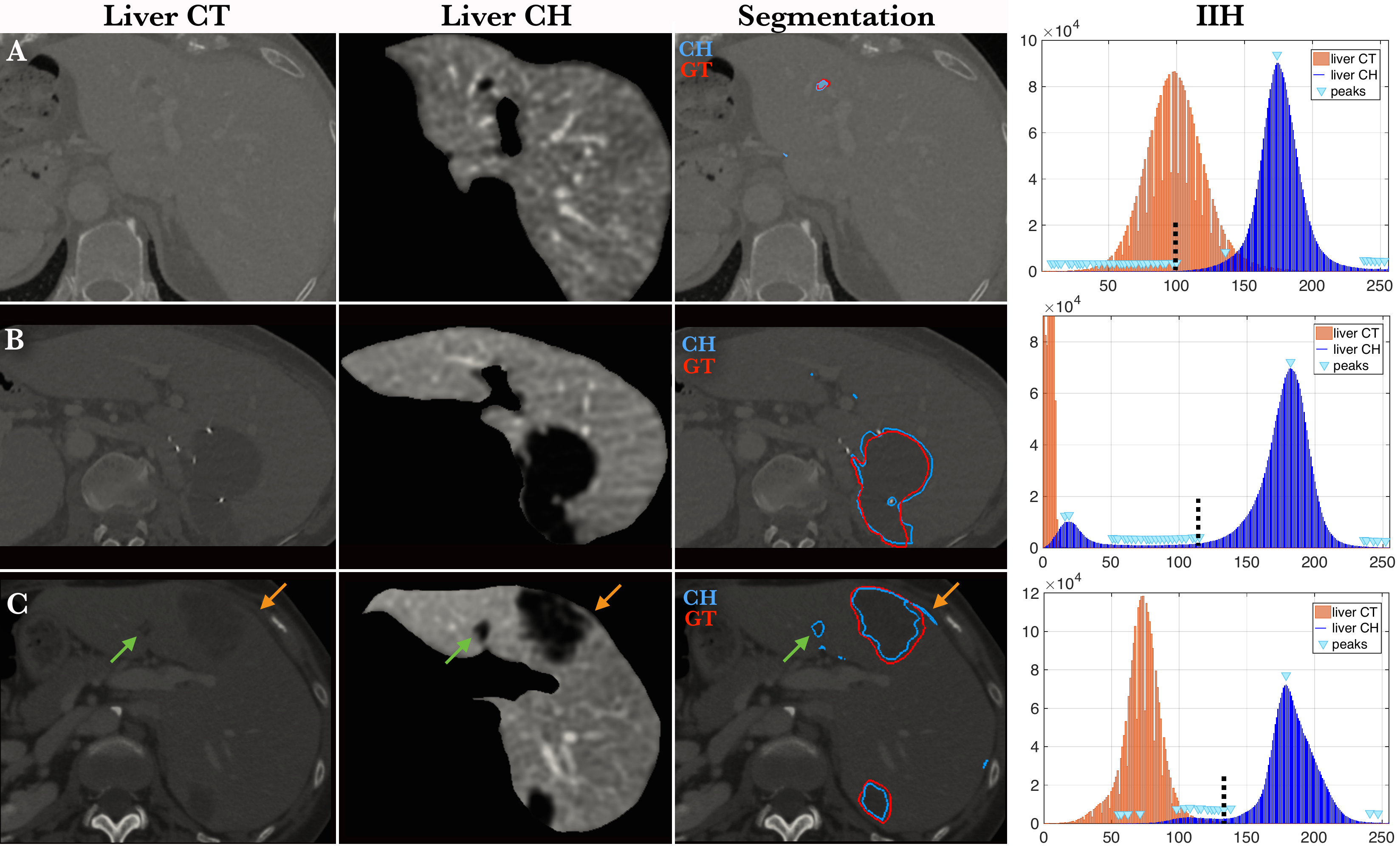}
\caption{Qualitative results showing original liver CT scan, liver after Cahn-Hilliard (CH) separation and comparison of the ground truth (GT) (red) and CH (blue) segmentation for three cases: A) case with small lesion, B) case with metal implants, C) case with segmentations artefacts at the liver border (orange arrow) and at region of liver folding (green arrow). Last column shows image intensity histogram (IIH) of the normalised liver CT before (orange) and after separation (blue). Cyan triangles indicate detected peaks, while the liver-lesion separation is marked by the dashed line.} 
\label{fig:Cases}
\end{figure}
%
%
%
\subsection{Qualitative results}
Figure \ref{fig:Cases} A) shows the capability of the method to enhance and detect small lesions. In this case the lesion mode in the IIH is less pronounced, nevertheless the correct separation is still detected. Figure \ref{fig:Cases} B) shows a liver volume with metal implants. Since CHS depends only on difference in image intensities, not the absolute values, presence of the metal artefacts does not influence the segmentation. Furthermore, using the metal implants as landmarks, it can be seen that the liver-lesion interface is preserved. However, segmentation based only on intensity thresholding can not distinguish between lesions and other artefacts with similar intensities, which might appear at the liver border or in regions of liver folding (Fig. \ref{fig:Cases} C). Furthermore, the method is not able to detect very small lesions in low resolution data, i.e. small lesions that appear only in 1-2 slices. 
\subsection{Quantitative results}
The CHS method was tested on the hypointense lesion from the LITS training set 2. For comparison purposes, the set was divided into two groups: 1) 3Dircadb dataset and 2) the rest of the set, referred to as LITS-hypo. Table \ref{table:Results} shows results of the CHS method in comparison with other automatic methods. High detection rate illustrates the capability of the method to enhance and separate lesions. On the 3Dircadb dataset, the CHS method performed better than the convolutional neural networks \cite{Christ:2017b}. The Dice scores on LITS-hypo test are lower than on 3Dircadb set for two reasons. First, the set contains several very small lesions present only in 1-2 slices. Second, the CHS method depends on the quality of the liver segmentation. Liver foldings and shadows at the liver borders tends to increase the number of false positives. A comparison with other methods on LITS training set is currently not possible, however this set helped to identify weak points of the CHS method. These weaknesses could be addressed by applying a classifier trained to distinguish between lesions and other artefacts.
%
%
%
\begin{savenotes}
\begin{table}[!]
\caption{Quantitative results of automatic liver lesions segmentation methods.  Scores are reported as presented in the original papers.}
\centering
\label{table:Results}
\begin{tabular*}{\hsize}{@{\extracolsep{\fill}} llccccc}
\hline
\hline
Approach & Dataset & Dice  &     Sensitivity  & Specificity  & Precision & Detection  \cr
\hline
\hline
CHS & 3Dircadb & $0.61\pm 0.22$ & $0.64\pm 0.18$ & $0.99\pm 0.01$ & $0.65\pm 0.27$ & $0.73\pm 0.25$ \cr
CHS & LITS-hypo & $0.53 \pm 0.27$ & $0.70 \pm 0.21$ & $0.98 \pm 0.02$ & $0.52 \pm 0.30$ & $0.85 \pm 0.20$ \cr 
Christ \cite{Christ:2017b}  & 3Dircadb & $0.56 \pm 0.27$    &   -     &  -      &  -     & -   \cr
Schweir \cite{Schwier:2011} & private & - & - & - & 0.53  &0.77  \cr
Massoptier \cite{Massoptier:2008} & private & - & 0.82 & 0.87 & - & - \cr
\hline
\end{tabular*}
\end{table}
\end{savenotes}
%
%
\section{Conclusion}
We have presented a novel automated and unsupervised method for segmentation of lesions in liver CT scans. The ability of the CHS method to enhance and detect lesions, allows to reach state-of-the-art results with simple thresholding of the Chan-Hilliard solution. We expect that combining the CHS with more discriminative learning approaches will enhance the quality of the segmentations. Application of the CHS method is not limited only to liver lesions and similar structures as lesions in spleen or ultrasound images. By modification of the chemical potential, the CHS method can be used for separation of multiple phases,  making it a promising tool for image preprocessing.

\end{document}